\documentclass[11pt]{article}

\usepackage[preprint]{acl}

\usepackage{times}
\usepackage{latexsym}
\usepackage{booktabs}

\usepackage[T1]{fontenc}

\usepackage[utf8]{inputenc}

\usepackage{microtype}

\usepackage{inconsolata}

\usepackage{pifont}    
\usepackage{caption}
\usepackage{tcolorbox} 
\usepackage{graphicx}  
\usepackage{todonotes}
\usepackage{xcolor}
\usepackage{multirow}
\usepackage{url}
\usepackage[acronym]{glossaries}
\usepackage{cleveref}
\usepackage{subcaption}

\newcommand{\cmark}{\ding{51}} 
\newcommand{\xmark}{\ding{55}} 
\newcommand{\qmark}{?}         
\usepackage{longtable}
\usepackage{booktabs}
\usepackage{array}
\usepackage{fontenc}   
\usepackage[utf8]{inputenc}
\usepackage[T2A,T1]{fontenc}
\usepackage{CJKutf8}

\newcommand{\cyr}[1]{{\fontencoding{T2A}\selectfont #1}}
\newcommand{\zh}[1]{\begin{CJK}{UTF8}{bsmi}#1\end{CJK}}

\newcommand{\cl}{CommonLID}

\glsdisablehyper
\newacronym{fpr}{FPR}{false positive rate}
\newacronym{lid}{LID}{language identification}
\newacronym{llm}{LLM}{large language model}

\title{CommonLID: Re-evaluating State-of-the-Art Language Identification Performance on Web Data}

\author{
\bf
 Pedro Ortiz Suarez$^{1,*,\dag}$,
 Laurie Burchell$^{1,*}$,
 Catherine Arnett$^{2,*}$,
\\
\bf
 Rafael Mosquera-Gómez$^{3,4}$,
 Sara Hincapie-Monsalve$^{3,4}$,
 Thom Vaughan$^{1}$,
\\
\bf
 Damian Stewart$^{1}$,
 Malte Ostendorff$^{1}$,
 Idris Abdulmumin$^{5}$,
 Vukosi Marivate$^{5,6}$,
\\
\bf
 Shamsuddeen Hassan Muhammad$^{7}$,
 Atnafu Lambebo Tonja$^{8}$,
 Hend Al-Khalifa$^{9}$,
\\
\bf
 Nadia Ghezaiel Hammouda$^{10}$,
 Verrah Otiende$^{11}$,
 Tack Hwa Wong$^{12}$,
\\
\bf
 Jakhongir Saydaliev$^{13}$,
 Melika Nobakhtian$^{14,15}$,
 Muhammad Ravi Shulthan Habibi$^{16}$,
\\
\bf
 Chalamalasetti Kranti$^{17}$,
 Carol Muchemi$^{18}$,
 Khang Nguyen$^{19}$,
\\
\bf
 Faisal Muhammad Adam$^{20}$,
 Luis Frentzen Salim$^{21,22}$,
 Reem Alqifari$^{9}$,
 Cynthia Amol$^{23,24}$,
\\
\bf
 Joseph Marvin Imperial$^{25,26}$,
 Ilker Kesen$^{27}$,
 Ahmad Mustafid$^{28}$,
 Pavel Stepachev$^{29}$,
\\
\bf
 Leshem Choshen$^{30,31,32}$,
 David Anugraha$^{33}$,
 Hamada Nayel$^{34}$,
 Seid Muhie Yimam$^{35}$,
\\
\bf
 Vallerie Alexandra Putra$^{36}$,
 My Chiffon Nguyen$^{37}$,
 Azmine Toushik Wasi$^{38}$,
\\
\bf
 Gouthami Vadithya$^{39}$,
 Rob van der Goot$^{40}$,
 Lanwenn ar C'horr$^{41}$,
 Karan Dua$^{28}$,
\\
\bf
 Andrew Yates$^{42}$,
 Mithil Bangera$^{39}$,
 Yeshil Bangera$^{39}$,
 Hitesh Laxmichand Patel$^{43}$,
\\
\bf
 Shu Okabe$^{44}$,
 Fenal Ashokbhai Ilasariya$^{45}$,
 Dmitry Gaynullin$^{28}$,
 Genta Indra Winata$^{46}$,
\\
\bf
 Yiyuan Li$^{47}$,
 Juan Pablo Martínez$^{48,49}$,
 Amit Agarwal$^{50}$,
 Ikhlasul Akmal Hanif$^{51}$,
\\
\bf
 Raia Abu Ahmad$^{52}$,
 Esther Adenuga$^{53}$,
 Filbert Aurelian Tjiaranata$^{16}$,
\\
\bf
 Weerayut Buaphet$^{54}$,
 Michael Anugraha$^{28}$,
 Sowmya Vajjala$^{55}$,
 Benjamin Rice$^{56}$,
\\
\bf
 Azril Hafizi Amirudin$^{57}$,
 Jesujoba O. Alabi$^{58}$,
 Srikant Panda$^{59}$,
 Yassine Toughrai$^{60}$,
\\
\bf
 Bruhan Kyomuhendo$^{5}$,
 Daniel Ruffinelli$^{61,62}$,
 Akshata A$^{28}$,
 Manuel Goulão$^{63}$,
 Ej Zhou$^{64}$,
\\
\bf
 Ingrid Gabriela Franco Ramirez$^{28}$,
 Cristina Aggazzotti$^{42}$,
 Konstantin Dobler$^{65,66}$,
\\
\bf
 Jun Kevin$^{67}$,
 Quentin Pagès$^{28}$,
 Nicholas Andrews$^{42}$,
 Nuhu Ibrahim$^{68}$,
\\
\bf
 Mattes Ruckdeschel$^{40}$,
 Amr Keleg$^{51}$,
 Mike Zhang$^{27}$,
 Casper Muziri$^{5}$,
 Saron Samuel$^{42}$,
\\
\bf
 Sotaro Takeshita$^{61}$,
 Kun Kerdthaisong$^{69}$,
 Luca Foppiano$^{70,1}$,
 Rasul Dent$^{71}$,
\\
\bf
 Tommaso Green$^{61}$,
 Ahmad Mustapha Wali$^{72}$,
 Kamohelo Makaaka$^{5}$,
 Vicky Feliren$^{73}$
\\
\bf
 Inshirah Idris$^{74}$,
 Hande Celikkanat$^{1}$,
 Abdulhamid Abubakar$^{75}$,
 Jean Maillard$^{76}$,
\\
\bf
 Benoît Sagot$^{71}$,
 Thibault Clérice$^{71}$,
 Kenton Murray$^{42}$,
 Sarah Luger$^{4}$
\\
\\
 $^{1}$Common Crawl Foundation,
 $^{2}$EleutherAI,
 $^{3}$Factored AI,
 $^{4}$MLCommons
 \\
$^{*}$Equal contribution. See Appendix \ref{app:affiliations} for all affiliations.
}

\begin{document}

\maketitle


\begin{abstract}
\Gls{lid} is a fundamental step in curating multilingual corpora. However, \gls{lid} models still perform poorly for many languages, especially on the noisy and heterogeneous web data often used to train multilingual language models. In this paper, we introduce CommonLID: a community-driven, human-annotated \gls{lid} benchmark for the web domain, covering 109 languages. Many of the included languages have been previously under-served, making CommonLID a key resource for developing more representative high-quality text corpora. We show CommonLID's value by using it, alongside five other common evaluation sets, to test eight popular \gls{lid} models. We analyse our results to situate our contribution and to provide an overview of the state of the art. In particular, we highlight that existing evaluations overestimate \gls{lid} accuracy for many languages in the web domain. We make CommonLID and the code used to create it available under an open, permissive license.
\end{abstract}

\renewcommand{\thefootnote}{\fnsymbol{footnote}}
\footnotetext[2]{Correspondence to \href{mailto:pedro@commoncrawl.org}{pedro@commoncrawl.org}}
\renewcommand{\thefootnote}{\arabic{footnote}}

\section{Introduction}



Language technologies should be useful tools for speakers of all languages, but this is far from the case. For example, even the most powerful language models only work well for a small number of languages (namely English and Mandarin Chinese). This is largely due to a lack of available high-quality datasets for many languages \citep{kreutzer-etal-2022-quality,foroutan2025conlid}. 

Language identification (\Gls{lid}) is one of the first steps in creating larger and better datasets.  Despite claims that \gls{lid} is ``solved'' (\citealp[e.g.][]{mcnamee_2005_solved}), \gls{lid} performance is still quite low for many languages including in key domains, such as web data. Even in English, \gls{lid} models perform worse on social media text and non-standard varieties \citep{blodgett-etal-2017-dataset}, which make up a large portion of web-crawl-based pre-training datasets such as those derived from Common Crawl\footnote{\url{https://commoncrawl.org/}}. 

In addition, many \gls{lid} systems either do not explicitly support under-served languages or perform poorly on them. For such languages, a majority of the available training data is religious text \citep{foroutan2025conlid}, meaning that \gls{lid} models covering these languages are expected to work on both domains and registers on which they were not trained. This limits practitioners' ability to make use of web data as a valuable potential source of training examples, since under most approaches, \gls{lid} systems will continue to perform poorly on web data for under-served languages. This in turn limits the development of language technologies such as large language models for languages other than English (LOTE), especially low-resource languages. Improving \gls{lid} models would allow us to improve datasets and in turn improve language model performance for many languages.

To help address these problems, we introduce \textbf{CommonLID}, a crowd-sourced \gls{lid} dataset, which we release under an open permissive license.\footnote{\url{https://huggingface.co/datasets/commoncrawl/CommonLID}.} The dataset is composed of line-level annotations by native speakers on web text coming from Common Crawl, created in collaboration with over eighty annotators.\footnote{All annotators who annotated more than 100 documents were invited to be co-authors on this paper.} The dataset includes 109 language varieties, 78 of which contain at least 100 lines. CommonLID is intended as a \gls{lid} benchmark to highlight where current datasets may be overestimating performance in low-resource, heterogeneous and noisy web contexts. This will help develop more robust models.

To facilitate future research in this topic, we also provide a comprehensive comparison of eight of the most common \gls{lid} models on \gls{lid} across multiple domains, including web data. We show that \gls{lid} is far from solved: using domain-specific evaluation datasets, we see that no existing model performs well across all domains. Developing better \gls{lid} systems is essential for reducing crosslingual inequity in the field. To do this, we need better domain-appropriate \gls{lid} evaluation datasets for more languages.

\section{Issues Affecting LID datasets}

\subsection{Training Datasets}

Issues with the datasets used to train \gls{lid} systems allows us to identify problems in the resulting systems. By identifying weaknesses in \gls{lid} systems, we can design targeted evaluations for the development of better models.

\paragraph{Lack of Open Data}
The datasets used to train the most popular \gls{lid} systems (e.g. CLD2 \citep{cld2}, CLD3 \citep{cld3}, FUN-LangID \citep{caswell2024funlangid}, fasttext \citep{joulin-etal-2017-bag}) are not publicly available. Other models (e.g. \citet{okorie2025fair}) do not release training data due to legal concerns. Whilst there are models which make their data available (e.g. Franc \citep{pyfranc}, GlotLID \citep{kargaran-etal-2023-glotlid}, and OpenLID \citep{burchell-etal-2023-open}), there is still a lack of open data for training SOTA \gls{lid} models, especially in the long tail of language coverage.

\paragraph{Language Coverage}
Many of the most popular \gls{lid} models do not support more than approximately 200 languages, reflecting the coverage of most existing multilingual datasets: e.g. Universal Dependencies \citep{nivre-etal-2016-universal,nivre-etal-2020-universal} and Europarl \citep{koehn-2005-europarl, tiedemann2012opus}. Newer corpora have been released aimed at combatting this. One example is The Corpus of Global Language Use \citep{dunn2020corpus}, which provides increased representation to regions such as South Asia, Sub-Saharan Africa, and Oceania.
Smol \citep{caswell-etal-2025-smol} is another recent effort, covering 123 languages which are rarely represented in other datasets. There have also been efforts to develop region-specific datasets, such as AfroLID (\citealp[]{adebara-etal-2022-afrolid}; African languages) and  \citet{blaschke-etal-2023-survey} (Germanic languages). This can offer targeted improvements for languages in these regions.

\paragraph{Domain and Register}
For \gls{lid} systems that cover over a few hundred languages, the primary source of data for most low-resource languages comes from religious texts: for example, the Parallel Bible Corpus \citep{mayer-cysouw-2014-creating} which covers over 1000 languages, and MIL-TALE \citep{brown-2014-non} which covers 2110 languages with the data sourced largely from religious text. Since religious data is all that is available for training for many languages in the so-called ``long tail'', most of the languages covered by a high-coverage \gls{lid} model will only have been trained on religious text, limiting domain generalisation for these languages \citep{costa2022no,goot-2025-identifying}. For example, GlotLID has been shown to have a bias towards selecting religious text when classifying heterogeneous web data \citep{penedo2025fineweb}. 

Many existing corpora do not include any web data, usually due to its noisy nature (e.g. GlotLID, OpenLID). A notable exception is Smol, which is comprised of professionally translated web data. Instead, many datasets are derived from ``cleaner'' sources like Wikipedia (e.g. FLORES; \citealp{nllb-24}) and government documents (Europarl, UDHR), which are distinct from the web domain in style and register.

\paragraph{Human Verification}
The language labels of many datasets used for training \gls{lid} models are not human verified, leading to questions around their reliability. For example, two of the largest multilingual datasets in OPUS \citep{tiedemann2012opus}, CCMatrix \citep{schwenk-etal-2021-ccmatrix}) and the Corpus of Global Language Use \citep{dunn2020corpus} were not manually checked. Later work by \citet{kreutzer-etal-2022-quality} found that the automatically-assigned labels in many large-scale datasets are inaccurate, which has a deleterious effect on downstream models.

\subsection{Evaluation Datasets}

Evaluation sets have their own specific issues in addition to those affecting training data more broadly, which  motivate the development of CommonLID.

\paragraph{Web Domain and Noisy Data}
Most systems are evaluated on clean, high-quality datasets such as FLORES and UDHR \citep{costa2022no,kargaran-etal-2023-glotlid, burchell-etal-2023-open}. 
Such datasets are professionally translated and represent non-fiction and legal domains. They do not include many of the artifacts of web data, and therefore may not be indicative of performance on such text.

\paragraph{Decreasing Availability}
Some datasets are not publicly available, such as the AfroLID test split, or were available but are no longer allowed to be used, such as JW300 \citep{agic-vulic-2019-jw300} and Twitter \citep{zubiaga2016tweetlid}. These cases are part of a growing trend in which permissions for data usage are being retracted, and is even more common for  higher-quality and more frequently used data sources \citep{longpre_2024_consent}. Consequently, it is increasingly important to create and use open evaluation datasets to support replicable research.

\paragraph{Dataset Size}
Some key datasets only contain a small number of items per language. UDHR, for example, only contains approximately 90 lines per language. Bible data are an exception, as they are large enough, are available and offer excellent language coverage. However, such data is from a limited domain and is often included as training data, especially for under-served languages.

\section{Annotation Collection/Dataset Creation}\label{sec:dataset_creation}


Collecting annotations for the CommonLID dataset was a multi-step, community driven endeavour. The process had three main parts. Firstly, we collected multilingual web data by sampling from recent filtered Common Crawl crawls and MADLAD-400 \citep{kudugunta-etal-2023-madlad}. Secondly, we recruited annotators, primarily from the NLP community. Thirdly, we collected annotations via a custom interface, often via hosted hackathons. We discuss the annotation process in more detail below.




\paragraph{Sampling Common Crawl}
We sampled data for contributors to annotate using the Ungoliant pipeline \citep{abadji-etal-2021-ungoliant} from the OSCAR project \citep{abadji-etal-2022-towards,ortiz-suarez-etal-2019-asynchronous}. We use three \gls{lid} models for selection: fastText \citep{joulin-etal-2017-bag,joulin-etal-2016-fasttextzip}, OpenLID \citep{burchell-etal-2023-open} and GlotLID \citep{kargaran-etal-2023-glotlid}.

Samples are taken from the WET files (text extracted from HTML) of the \texttt{CC-MAIN-2024-22} and \texttt{CC-MAIN-2025-05} crawls. We sampled up to 1,000 documents per language from both crawls with each of the three \gls{lid} models, meaning that a maximum of 6,000 documents were sampled per language (2 crawls $\times$ 1000 documents $\times$ 3 \gls{lid} models). We used the OSCAR automatic text quality annotations introduced in \citet{abadji-etal-2022-towards} to filter out short and noisy documents, since they are based on very simple heuristics that do not target any specific domain or language register.

We also included the noisy splits from MADLAD-400 to create data for annotation in order to increase the amount of data available for some less-represented languages, and because it uses a distinct architecture for \gls{lid} in web data \citep{caswell-etal-2020-language}. We sampled 1000 documents by language from the clean and noisy splits of both the original version and version 1.5, resulting in 4000 documents per language.


We note that selecting samples using existing \gls{lid} systems is a key limitation of our methodology because it means that only data which is already recognised by a \gls{lid} system will be selected. It also biases the data to genres and language registers which match the models' training data. However, given the scale of web data and the language skew on the web, it would have been unrealistic to ask annotators to check a random sample of web data until they found content in their language, especially for under-represented languages.\footnote{\url{https://commoncrawl.github.io/cc-crawl-statistics/plots/languages}} Pre-annotating data with existing models is thus a necessary compromise to achieve better language representation.


\paragraph{Recruitment}

We invited members of the NLP community to participate in the creation of CommonLID, since researchers who speak and work with particular languages are motivated to help improve \gls{lid} models for their languages of interest.

We recruited via a wide range of social media and NLP Discord servers. We also partnered with grass-roots NLP organisations, which focus on driving NLP progress for languages of a particular region. This helped us connect with researchers at different career stages, especially early-career and aspiring researchers, around the world. In collaboration with these grass-roots organisations, we held virtual hackathon events to create a space for annotators to ask questions, get feedback, interact with each other, and set aside time for annotations. 

Since contributors were mostly members of the NLP community, we found that they understood the value of annotated data in their languages. This meant that we had good engagement from the community and higher-quality annotations than we might expect from crowd-sourced workers \citep{fort-etal-2011-last,fort-etal-2014-crowdsourcing}.



\paragraph{Annotation Platform}

We modified the latest version\footnote{\url{https://github.com/mlcommons/dynabench/}} of Dynabench \citep{kiela-etal-2021-dynabench} for the annotation interface. In this interface, users first register in order to track their number of annotations. When they start annotating, they choose the language they want to validate data for and are then presented with a series of plain text documents from the pre-annotated samples in their language of choice. Annotators highlight the sections of the documents that are written in their language of choice and are also instructed to annotate other sections of the document that might be written in other languages (provided they recognize them).

Users are encouraged to highlight entire lines if possible: that is, they should do line-level language identification. The interface also allowed users to highlight or do annotations at the character-level, so that they can complete annotations when the language of a complete line was impossible to determine. Large documents were truncated in order for the interface to be responsive. Screenshots of the interface, as well as the complete annotations guidelines can be found in \Cref{app:annotation_platform}.


\paragraph{Incentives for Participation}

All contributors who annotated at least 100 documents were offered authorship on the dataset paper. We also offered authorship to contributors all available documents if we had fewer than 100 for that language, which was the case for some of the most low-resource languages. Offering authorship was a key design decision of this project, consistent with the principles of participatory design proposed in \citet{caselli-etal-2021-guiding}. In particular, we wanted to engage in ethical community collaboration: treating community members with respect, equity and reciprocity. Authorship represents an important way to recognize the intellectual contributions of  annotators and increases equity between all contributors of the project. 
This reflects the goal of improving language representation in NLP datasets---and, as a result, models. But we believe in order to do this responsibly, this requires deep cooperation with language communities and giving contributors ownership over their data. 

We also incentivised contributors to annotate more documents through a leaderboard displayed on the annotation platform. Each contributor's username was displayed along with the number of documents they annotated. While most contributors annotated between 100 and 200 documents, some participants were far more prolific: 10 annotated over 1,000 documents, and 5 contributors who annotated over 2,000 documents.





\section{CommonLID dataset}

We finished collecting annotations for the version of \cl{} described in this paper on 25 November 2025. We then processed the collected data to create a corpus annotated with the dominant language variety in each line.


\subsection{Dataset preparation}\label{sec:dataset_preparation}

\paragraph{Creating line-level labels} 
Reformatting the annotated data to be labelled at the line level was not straightforward. Firstly, some participants had misunderstood the instructions and either labelled the text at the word level, resulting in multiple labels per line, or labelled all text in the page as one language when in fact multiple languages were present. Secondly, accurate multilingual sentence splitting is not a trivial task. Thirdly, some lines contain content which is hard to classify into any particular language variety (e.g. URLs, proper nouns).

To avoid complications with multilingual sentence splitting, we simply split the text based on new lines. We then removed leading and trailing white space and deduplicated language variety label and line pairs. We used Levenshtein distance between lines to identify lines which were extremely similar but had been assigned different language variety labels to allow for fuzzy splitting. 

We noticed many lines with multiple labels were very short to the extent that \gls{lid} was not a meaningful task, so we filtered out any lines with fewer than 10 characters. We also noticed that many non-English languages contained a significant number of English lines. We therefore used the \texttt{langid.py} \gls{lid} tool\footnote{\url{https://github.com/saffsd/langid.py}} to predict whether lines were English, as a fast and simple \gls{lid} tool. Any short lines predicted to be English or longer lines with predicted probability of being English $ >0.7 $ were filtered out, based on empirical results. We exempted lines labelled as Scottish Gaelic, Irish Gaelic, Nigerian Pidgin, Southern Sotho, and Shona from the English filter since the false positive rate was too high. 

\paragraph{Reconciling multiple labels} After the initial cleaning, one of the authors (a British English speaker able to read Latin and Arabic scripts) manually inspected all rows with multiple labels. Some lines remained which were actually English but labelled as different distinct languages, or conversely labelled as English but were not in reality. We assumed this was an artifact of over-use of the `select all' feature in the interface (see \Cref{fig:platform_screenshot}, top right) and removed these lines. Short lines containing solely non-linguistic content (e.g. numbers, URLs, emoji) were also removed. 

We found two categories of annotation where we could not reconcile the distinct labels. The first category was lines labelled with both the macro- and micro-language label (e.g. Arabic and Egyptian Arabic, Malay and Indonesian). The second category was lines containing multiple languages but without a dominant language. In both cases, we kept multiple copies of the line with the different annotations. The existence of multiple valid labels for a single line is a key limitation of our labelling scheme and should be addressed in future work \citep{keleg-etal-2023-aldi,burchell-etal-2024-code}. 


\paragraph{Per-language audit}
For each language class, we checked a random sample of approximately 100 lines to check for language correctness or other quality issues, following the heuristics in \citet{burchell-etal-2023-open}. Three language variety classes were spurious, only containing 1-2 lines which were clearly not in the intended language (\texttt{gux}, \texttt{swe}, and \texttt{tag}). These were removed.

\subsection{Inter-annotator agreement}
\label{sec:iaa}

Due to our reliance on volunteer annotators and our focus on under-served languages, only 12.9\% of lines in the dataset were labelled by more than one annotator.  Nevertheless, we explore this subset of the data to estimate inter-annotator agreement. The final version of CommonLID is deduplicated at the tag/line level, so we calculate statistics from the dataset after filtering short spans but prior to applying other quality filters described in \Cref{sec:dataset_preparation}.

Of the 67,625 lines in the dataset with multiple annotations (out of 523,154 overall), 2,137 were annotated with multiple differing tags (3.2\%). We inspected these lines to assess whether these were truly difficult examples or simple mistakes.

We found that all the 16 lines with four or more differing tags were boilerplate, presumably `caught' when the participant was labelling text in a different language (e.g. ``Post a Comment'' was labelled as `ind', `mal', `vie', `swh', and `eng'). Of the 78 lines assigned three different tags, 26 were English-language boilerplate (e.g. ``Related Articles'' was labelled as `ind', `vie', and `eng') and the remaining 52 lines were Arabic text which had been labelled as Modern Standard Arabic, Arabic macrolanguage, and an Arabic dialect (`arb', `ara', and `arz'/`ars' etc.). 

As for the 2043 lines with two differing annotated labels, over half were annotated as Modern Standard Arabic/Arabic macrolanguage (`arb'/`ara', 746 lines) or Modern Standard Arabic/Egyptian Arabic (`arb'/`arz', 283 lines). Overall, 1218 lines with two tags were annotated as two kinds of Arabic. The nature of Arabic as a language continuum means that it is unsurprising that human annotators disagree on how to assign discrete language labels. 658 of the remaining lines with two labels were labelled as English and another language, which was unsurprising given the dominance of English on the web. Finally, we found 254 examples of two distinct tags where neither tag was English or an Arabic language. These were mostly macro/micro language labelling (e.g. `cmn'/`zho', 77 lines), close but distinct languages (e.g. `lav'/`lvs', 19 lines) or distant but historically-linked languages (e.g. `gug'/`spa', 34 lines).

If we filter mis-labelled English boilerplate, this leaves 1,524 lines which could be counted as actual annotator disagreement: 2.3\% of the multiply-annotated lines. We feel this is reasonable, given the complications of discrete language labelling.

\subsection{Dataset description}

\begin{figure}[ht!]
    \centering
    \includegraphics[trim=45 140 45 50,clip,width=\linewidth]{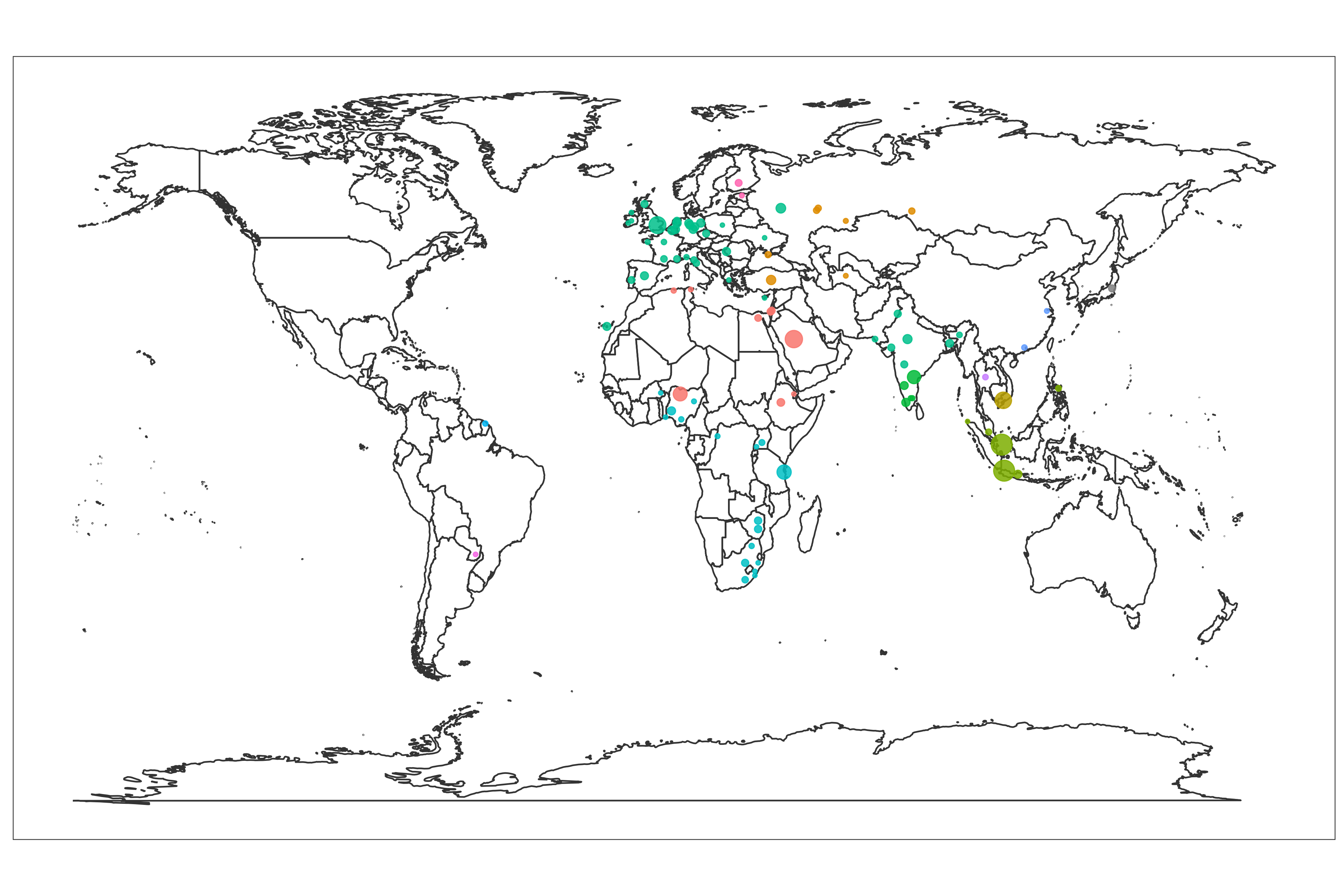}
    \caption{Illustrative geographical distribution of the language varieties in \cl{}. The dot size represents the number of annotated lines, and colour represents language family.}
    \label{fig:wmdqs_map}
\end{figure}

\cl{} contains 373,230 lines in total with a mean line length of 215.5 characters. This makes \cl{} over ten times larger than Smol \citep{caswell-etal-2025-smol}. There are 109 language varieties, with the largest containing 43,189 lines (\texttt{uzb}), and the 4 smallest containing a single line (\texttt{ace}, \texttt{pol}, \texttt{grc}, \texttt{gle}). The macro-average number of lines per class is 3424.1, and 78 classes contain more than 100 examples. \Cref{fig:wmdqs_map} illustrates the rough geographical distribution of the languages covered in \cl{}, with the dot size representing the number of lines. \Cref{sec:appendix_stats} contains a full breakdown of number of lines and mean line length per language class. 

\section{Evaluation}
We analysed the performance of eight widely-used existing \gls{lid} models on our dataset, and compared the performance of the same models on comparable existing evaluation datasets. This extensive testing situates CommonLID in the context of previous work and demonstrates its value as a web-derived human-annotated evaluation dataset. It also serves as a reference for the state of the art for \gls{lid} models to aid future research and development. We release all our raw evaluation scores and analysis code.\footnote{\url{https://github.com/commoncrawl/commonlid-eval}.}

\subsection{Evaluation datasets}

The majority of our evaluation datasets are openly available which is our preference. However, we also use two restricted datasets for where we could not find an open alternative: a Bible dataset since it includes an extremely wide range of languages, and a social media dataset since it includes colloquial text often found on the web. We provide further details about all evaluation datasets in \Cref{app:eval_datasets}. 

Three of the datasets (SmolSent, Bibles, and social media) had a large variation in the number of examples per class in their original form.
This had the double effect of greatly increasing inference times due to the large classes as well as leading to unrepresentative results for the smallest classes. To combat this, we discarded classes in these datasets with fewer than 300 examples, then sampled 300 lines at random from the remaining classes to form our test splits for these datasets.

\begin{table*}[ht!]
    \centering
    \small
    \begin{tabular}{lrrrrrrrrrrrr}
        \toprule
        &\multicolumn{2}{c}{FLORES+} & \multicolumn{2}{c}{SmolSent} & \multicolumn{2}{c}{UDHR-LID}  & \multicolumn{2}{c}{Bibles} & \multicolumn{2}{c}{Social Media} & \multicolumn{2}{c}{CommonLID} \\
        &\multicolumn{2}{c}{209 languages} & \multicolumn{2}{c}{82 languages} & \multicolumn{2}{c}{418 languages}  & \multicolumn{2}{c}{1047 languages} & \multicolumn{2}{c}{66 languages} & \multicolumn{2}{c}{109 languages} \\
        \cmidrule(lr){2-3} \cmidrule(lr){4-5} \cmidrule(lr){6-7} \cmidrule(lr){8-9} \cmidrule(lr){10-11} \cmidrule(lr){12-13} 
        & all & cov. & all & cov. & all & cov. & all & cov. & all & cov. & all & cov. \\
        \midrule
AL & 15.9 & 67.9 {\tiny{(49)}} & 42.3 & 69.4 \tiny{(50)} & 15.2 & 67.6 \tiny{(94)} & 12.1 & 69.8 \tiny{(181)} & 1.5 & \textbf{98.0} \tiny{(1)} & 9.2 & 43.5 \tiny{(23)} \\ 
C2 & 45.0 & 87.8 \tiny{(105)} & 40.6 & 83.2 \tiny{(40)} & 24.0 & 81.7 \tiny{(121)} & 4.4 & 73.8 \tiny{(62)} & 76.8 & 85.9 \tiny{(59)} & 49.5 & \textbf{79.3} \tiny{(68)} \\ 
FT & 80.3 & 92.2 \tiny{(182)} & 48.1 & 83.8 \tiny{(47)} & 27.0 & 68.2 \tiny{(166)} & 3.3 & 43.8 \tiny{(79)} & \textbf{83.6} & 84.9 \tiny{(65)} & 49.3 & 72.6 \tiny{(74)} \\ 
FL & 68.8 & 86.2 \tiny{(165)} & \textbf{75.9} & 79.8 \tiny{(78)} & 59.0 & 78.2 \tiny{(314)} & 71.3 & 89.1 \tiny{(837)} & 58.7 & 65.5 \tiny{(58)} & 46.4 & 57.8 \tiny{(86)} \\ 
Fr & 61.9 & 80.4 \tiny{(160)} & 46.4 & 76.0 \tiny{(50)} & \textbf{93.4} & \textbf{95.9} \tiny{(407)} & 7.6 & 52.1 \tiny{(152)} & 62.4 & 66.4 \tiny{(62)} & 39.3 & 61.3 \tiny{(70)} \\ 
C3 & 28.1 & 71.5 \tiny{(78)} & 11.9 & 57.4 \tiny{(17)} & 11.0 & 54.4 \tiny{(80)} & 2.1 & 42.8 \tiny{(47)} & 63.1 & 75.7 \tiny{(54)} & 34.3 & 66.2 \tiny{(55)} \\ 
GL & \textbf{94.2} & \textbf{96.5} \tiny{(204)} & 74.7 & \textbf{91.4} \tiny{(67)} & 73.7 & 84.4 \tiny{(366)} & \textbf{82.3} & \textbf{93.0} \tiny{(927)} & 80.6 & 85.8 \tiny{(62)} & \textbf{60.4} & 68.6 \tiny{(96)} \\ 
OL & 83.5 & 90.4 \tiny{(193)} & 45.4 & 82.7 \tiny{(45)} & 25.1 & 67.3 \tiny{(156)} & 3.3 & 44.6 \tiny{(78)} & 83.4 & 88.7 \tiny{(62)} & 47.4 & 68.0 \tiny{(76)} \\
        \bottomrule
    \end{tabular}
    \caption{Macro-averaged F1 scores achieved by tested models on the evaluation sets: AfroLID (AL), CLD2 (C2), fasttext (FT), FUN-LangID (FL), pyFranc (Fr), CLD3 (C3), GlotLID (GL), and OpenLID-v2 (OL). Scores are calculated over the whole dataset (\textit{all}) and on the subset of language varieties covered by the model (\textit{cov.}). Count of languages in the evaluation set covered by the model in parentheses, highest score per column in \textbf{bold}.}
    \label{tab:results_data_x_model_f1}
\end{table*}

\subsection{LID models}


We test eight \gls{lid} models using the evaluation sets: AfroLID \citep{adebara-etal-2022-afrolid}, CLD2 \citep{cld2}, (NLLB) fasttext \cite{costa2022no}, FUN-LangID \cite{caswell2024funlangid}, pyFranc \cite{pyfranc}, CLD3 \citep{cld3}, GlotLID v4 \citep{kargaran-etal-2023-glotlid}, and OpenLID-v2 \citep{burchell-etal-2023-open}. We summarise details of these models in \Cref{tab:eval_models} in Appendix \ref{app:model_details}. 
Prior to prediction, we normalise the input text by removing surrounding white space, applying lowercasing, removing non-word characters and digits, and squeezing white space. We found this improved the robustness of \gls{lid} models to non-standard input without assuming an input language.

\subsection{Large language models}\label{subsec:llms}

Furthermore, we evaluate large language models (LLMs) from OpenAI as baselines. Specifically, we compare GlotLID with GPT-4o, GPT-4o-mini, GPT-5, GPT-5-mini in a zero-shot setting.
The corresponding LLM prompts are implemented with DSPy~\citep{khattab2024dspy} without optimisation.

\subsection{Label normalisation}\label{subsec:label_normalisation}
The \gls{lid} datasets and models tested in this work do not share a common language coding schema, meaning that the labels must be normalised to allow large-scale comparison. Our process is as follows:
\begin{enumerate}
\itemsep0em 
    \item Raw language code strings are trimmed at the first \texttt{`-'} or \texttt{`\_'} character to keep the leading segment (e.g., \texttt{en-US} → \texttt{en}, \texttt{zh\_Hant} → \texttt{zh}).
    \item The trimmed segment is parsed by the Python \texttt{iso639-lang} library\footnote{\url{https://github.com/LBeaudoux/iso639}, v2.5.1}, which provides comprehensive error reporting on invalid and deprecated language codes.
    \item The resulting language code is checked for ISO 639-3 compliance, mapping deprecated codes to their modern equivalents where possible. Dataset entries and model outputs which cannot be resolved to a single ISO 639-3 language are treated as undefined and discarded.
\end{enumerate}
Automatic normalisation in this way has the potential to miss labels which could be compared (e.g. macro- and micro- language codes). However, manual reconciliation of so many language labelling schemes would be infeasible.



\subsection{Metrics and scoring}
\label{sec:metrics_scoring}
Following previous work \citep{burchell-etal-2023-open,nllb-24,kargaran-etal-2023-glotlid}, we use F1 score and \gls{fpr} as our primary metrics. All calculated averages are macro-averages so that less-resourced language varieties are given the same weight as those with more examples in the evaluation sets.

\subsection{Comparing models}
\label{sec:comparing_models}
Comparing the performance of different \gls{lid} systems involves comparing between classifiers which output non-overlapping classes. Different \gls{lid} models cover different language varieties and cannot output the correct label for unseen languages. This means that when taking a simple average over the test set, models with more coverage have an advantage since a model will always score zero for a language it does not include.

That said, higher-coverage models are not always better in practice: low \gls{lid} accuracy for less-resourced languages results in unacceptably low quality labelled data, hindering performance downstream \citep{caswell-etal-2020-language}. A more reliable but lower-coverage \gls{lid} model may be preferable depending on the application.

\section{Results}
\label{sec:results}

\subsection{LID models} \label{ssec:lid-results}

\Cref{tab:results_data_x_model_f1} shows F1 scores achieved by each model on each dataset, macro-averaged over language varieties. Two scores are given: one calculated over the whole evaluation set where languages not included in the model score zero (\textit{all}), and one calculated over the subset of language varieties covered by the model (\textit{cov.}). The number of languages over which the macro-average is calculated in the latter case is also given. For all models, the difference in these two scores highlights the ongoing issue of determining a fair way of comparing \gls{lid} models with disparate language coverage. It also provides a fuller picture of the trade-off between coverage and specialisation, which is most apparent for AfroLID.

The results for the CommonLID dataset in \Cref{tab:results_data_x_model_f1} show that it is a challenging dataset, with most models only achieving F1 scores in the 60s averaged over the languages they cover. To illustrate some of the failure models, we present some examples of errors made by GlotLID on CommonLID in Appendix \ref{sec:examples_errors}. We believe the current low scores on CommonLID demonstrate its worth as a novel evaluation dataset, since it covers a different domain compared to other datasets available for under-served languages.

Some of the higher scores for other datasets can be attributed to a likely overlap between the model's training data and the evaluation set: Franc is trained on UDHR, GlotLID uses Bible data as a majority of its training data for many languages in the long tail, and the developers of FUN-LangID were also part of the team behind SmolSent (though the exact training data for both is unknown). The effect on reported scores is exacerbated for high-coverage models, since for these most language classes are in the long tail where training and evaluation data are more similar. Taking the macro-average weights all languages equally, so given the majority of the languages are in the long tail, the average overall score is high. These high scores do not necessarily reflect real-world performance, showing the need for more diverse, independent evaluation data for long-tail languages.

\begin{figure}[thb]
    \centering
    \includegraphics[width=0.47\textwidth]{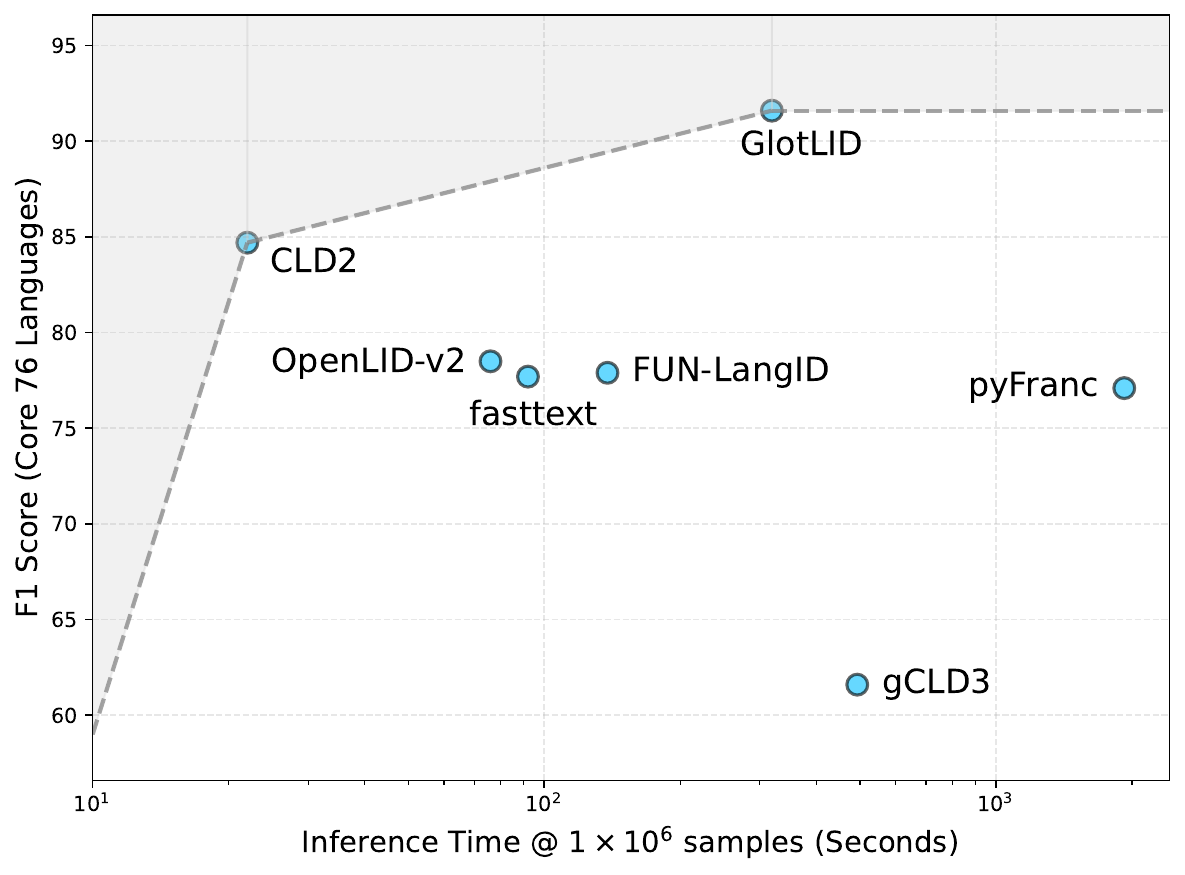}
    \caption{Inference speed vs. F1 (76 core languages, combined datasets)}
    \label{fig:frontier}
\end{figure}

\begin{table}[tbh]
        \centering
        \small
        \begin{tabular}{lrrrrrr}
            \toprule
            & \multicolumn{2}{c}{High-coverage} & \multicolumn{2}{c}{Afro*} & \multicolumn{2}{c}{Core} \\
            & \multicolumn{2}{c}{1351 languages} & \multicolumn{2}{c}{294 languages} & \multicolumn{2}{c}{76 languages} \\
            \cmidrule(lr){2-3} \cmidrule(lr){4-5} \cmidrule(lr){6-7}
             & F1 $\uparrow$ & FPR $\downarrow$ & F1 $\uparrow$ & FPR $\downarrow$ & F1 $\uparrow$ & FPR $\downarrow$ \\
            \midrule
            GL & 87.8 & 0.006\% & 88.7 & 0.02\% & 91.6 & 0.1\% \\
            FL & 76.3 & 0.02\% & 73.5 & 0.08\% & 77.9 & 0.2\% \\
            AL & --- & --- & 71.8 & 0.7\% & --- & --- \\
            Fr & --- & --- & --- & --- & 77.1 & 0.4\% \\
            FT & --- & --- & --- & --- & 77.7 & 0.5\% \\
            OL & --- & --- & --- & --- & 78.5 & 0.5\% \\
            C2 & --- & --- & --- & --- & 84.7 & 0.2\% \\
            C3 & --- & --- & --- & --- & 61.6 & 1.4\% \\
            \bottomrule
        \end{tabular}
        \caption{Performance for all test data combined for the subset of languages supported by all compared models. We give results for GlotLID (GL), FUN-LangID (FL), AfroLID (AL), pyFranc (Fr), fasttext (FT), OpenLID-v2 (OL), CLD2 (C2), and CLD3 (C3). Performance should only be compared within each column.}
        \label{tab:results_model_x_model_f1}
\end{table}

GlotLID shows strong performance for many evaluation sets and has the highest reported language coverage of the models we test. This comes at a cost of slower inference: \Cref{fig:frontier} plots inference speed versus F1 score for all models but AfroLID on a core set of 76 shared languages. CLD2 and GlotLID are on the Pareto frontier of compute-performance trade-off,  as GLotLID's improved performance comes at the cost of significantly slower inference. \Cref{tab:speed} contains full details of the inference speed for all models.

As mentioned in \Cref{sec:comparing_models}, it is often difficult to make a direct comparison between \gls{lid} models which cover different sets of language varieties. We decided that the fairest way to compare models directly was to evaluate them on test sets containing only the languages which were supported by all models. \Cref{tab:results_model_x_model_f1} shows the results for this type of evaluation. ``High-coverage'' compares results for GlotLID and FUN-LangID alone as the highest-coverage models, ``Afro*'' compares these models with AfroLID for coverage of the languages of Africa, and ``Core'' compares all models but AfroLID on a core set of 76 language varieties all share.\footnote{AfroLID's focus on the languages of Africa means that only one language, Afrikaans, is shared between all models.} Note that only metrics in the same column can be compared to each other. These results show that GlotLID performs strongly across all comparisons both in terms of F1 and \gls{fpr}.



\subsection{LLMs for LID} \label{sec:llm-results}

Given that LLMs are state-of-the-art for many NLP tasks, we report the results of four OpenAI GPT models based on three language groups in \Cref{tab:llm_results_model_x_model_f1}.
The language groups are the same as in \Cref{tab:results_model_x_model_f1}.
To reduce LLM inference costs, we down-sample the evaluation datasets (15k samples in total).
The best performing LID model, GlotLID, is used as a baseline.
Despite requiring magnitudes more  resources, the GPT models are outperformed by GlotLID.
The performance gap is smaller for the core languages (-1.8\% F1 with GPT-5)  and larger for the African languages (-30\% F1 with GPT-5).


\begin{table}[t]
        \centering
        \small
        \setlength{\tabcolsep}{4pt} 
        
        \begin{tabular}{lrrrrrr}
            \toprule
            & \multicolumn{2}{c}{High-cov.} & \multicolumn{2}{c}{Afro*} & \multicolumn{2}{c}{Core} \\
            & \multicolumn{2}{c}{1351 langs.} & \multicolumn{2}{c}{294 langs.} & \multicolumn{2}{c}{76 langs.} \\
            \cmidrule(lr){2-3} \cmidrule(lr){4-5} \cmidrule(lr){6-7}
             & F1 $\uparrow$ & FPR $\downarrow$ & F1 $\uparrow$ & FPR $\downarrow$  & F1 $\uparrow$ & FPR $\downarrow$  \\
            \midrule

            GlotLID & 90.0 & 0.01\% & 90.6 & 0.05\% & 93.5 & 0.13\% \\

            \midrule

            \multicolumn{7}{l}{\textit{OpenAI GPT models}} \\

            \rule{0pt}{3ex}4o-mini & 48.5 & 0.09\% & 43.6 & 0.69\% & 81.0 & 0.54\% \\
            
            4o & 62.0 & 0.07\% & 57.3 & 0.46\% & 89.0 & 0.33\% \\
            
            
            5-mini & 60.4 & 0.06\% & 55.1 & 0.43\% & 76.4 & 0.53\% \\

            5 & 70.3 & 0.05\% &  66.6 & 0.29\% & 91.8 & 0.22\% \\

            \bottomrule
        \end{tabular}
        \caption{A comparison of large language models from OpenAI's GPT family (4o-mini, 4o, 5-mini, 5) and GlotLID (best performining LID model). We report performance on the combined but down-sampled test data (15k test samples),  subsets as in \Cref{tab:results_model_x_model_f1}.}
        \label{tab:llm_results_model_x_model_f1}
\end{table}


\section{Discussion}\label{sec:discussion}

\paragraph{Creating CommonLID}

We created \cl{} in collaboration with many native speakers. The result significantly expands the availability of \gls{lid} web-domain evaluation data, especially for low-resource languages. Our work complements other collaborative efforts, like Smol, though unlike previous efforts we annotate texts originally in the target language rather than translated texts.


The creation of the dataset was constrained by the data available for annotation. We used existing LID models and datasets to select texts to annotate, meaning it was extremely difficult to collect texts in languages not supported by these models or datasets for contributors to annotate. During our hackathons, some participants were unable to annotate significant amounts of data because we had few documents available. This highlights one of the challenges of \gls{lid} model development: it is not simply a matter of gathering data. All existing tools and methods to \textit{begin} the annotation process break down for all but the highest resource languages. In the future, we hope to work with native speakers to collect textual resources in their languages to help improve language coverage in NLP technologies and as a step towards annotating such data.






\paragraph{Comparing \gls{lid} Models}

Given a range of \gls{lid} models, it is unclear how to compare them fairly. Each model has a different number of language labels it is trained to support. Unifying the language labels across models is far from straightforward, and even delineating boundaries between some language varieties is contentious \citep{burchell2024improving}. 

For many of the languages in the highest-coverage models like GlotLID and FUN-LangID, there is no test split distinct from the training data for many languages, since data for many languages in the long tail is already so sparse and usually limited to religious text. Therefore, it is currently impossible to evaluate \gls{lid} performance for the lowest resource languages in a meaningful way. This is one way in which current evaluation practices over-estimate \gls{lid} performance. 



\paragraph{State-of-the-Art \gls{lid}}
Considering the trade-off between coverage and inference speed, our results show that CLD2 and GlotLID perform best overall. CLD2 out-performs GlotLID in some contexts, but does not support lower-resource languages. GlotLID performs best on FLORES and Bibles, which represent some of the cleanest evaluation datasets. In particular, GlotLID's high F1 on Bible data reflects the fact that for most long-tail languages Bible data is the only  \gls{lid} training data for GlotLID. It is likely that performance for these languages would be more limited if the evaluation data covered a wider range of domains and were more distinct from the training data. 

That said, there is no clear top-performing model given the complexity of meaningful comparison and evaluation. What is clear is that \gls{lid}, especially in the long tail, remains challenging. There is no model that achieves >75\% F1 across all evaluation datasets, even when coverage is taken into account. There is a clear need for better \gls{lid} models. \cl{} helps to select and develop models that will work best on web data.

\section{Conclusion}
We presented \cl{}, a \gls{lid} evaluation dataset in the web domain. We detailed our highly collaborative data collection and annotation process, in which we worked with contributors from multiple language communities. To show \cl{} value as a benchmark, we used it to test eight popular \gls{lid} models alongside other common \gls{lid} evaluation datasets. We analysed our results to provide researchers with an overview of the state of the art in this task and thus support further work. Our data and code are made available to the community under an open, permissive license.


\section*{Limitations}

As mentioned in \Cref{sec:dataset_creation}, we used three existing models and one existing dataset in select samples to be validated and annotated by contributors. This inherently biases our dataset to the languages, genres and language registers supported by these models and dataset. Moreover, these three models were used in our evaluation, which could have potentially skewed the results. We aimed to mitigate this by evaluating CommonLID along with other existing datasets.

Other concerns are inherent to the task of language identification: for example, the existence of multiple labels for a given single line and the existence of macro- and micro- language. These can significantly complicate the annotation task, even for native speakers. We aimed to reduce the ambiguity by presenting full documents to the annotators so that they could have more context, but this does not remove all of the uncertainty inherent in the task. Moreover, as mentioned in \Cref{sec:comparing_models} and \Cref{sec:discussion}, evaluating and comparing models remains challenging, especially as we had to account for issues such as deprecated language codes, non-overlapping classes and existing data sparsity for many languages in the long tail. All of these factors complicate the evaluation and remain challenging to address.

We sourced the data used to build CommonLID from Common Crawl web crawl data. This means it has the potential to contain personally identifiable information or other harmful content. We mitigated this by using heuristic quality filters on the data prior to presenting it to participants and by making it easy for participants to contact us with any concerns. Moreover, we chose Common Crawl as a source because they have always respected robots.txt\footnote{\url{https://commoncrawl.org/ccbot}}, meaning that website owners have the option to opt out of crawling very easily.

Finally, given that we targeted annotations for a wide range of languages, we were unable to find more than one volunteer native speaker for many of them. This made it impossible to conduct a full inter-annotator agreement study for our data, though include analysis of the subset with multiple labels in \Cref{sec:iaa}.

We are not aware of any potential harms from this work. In fact, we hope this work helps to address harms from existing \gls{lid} models and helps improve cross-lingual equity in NLP.




\bibliography{custom}

\appendix
\section{Affiliations} \label{app:affiliations}

 $^{1}$Common Crawl Foundation,
 $^{2}$EleutherAI,
 $^{3}$Factored AI,
 $^{4}$MLCommons,
 $^{5}$University of Pretoria,
 $^{6}$Lelapa AI,
 $^{7}$Imperial College London,
 $^{8}$UCL,
 $^{9}$King Saud University,
 $^{10}$University of Hail,
 $^{11}$USIU-Africa,
 $^{12}$Universiti Teknologi PETRONAS,
 $^{13}$EPFL,
 $^{14}$Tehran Institute for Advanced Studies,
 $^{15}$Khatam University,
 $^{16}$Universitas Indonesia,
 $^{17}$University of Potsdam,
 $^{18}$Universität Trier,
 $^{19}$Michigan State University,
 $^{20}$NOUN (ACETEL),
 $^{21}$Academia Sinica,
 $^{22}$National Taiwan University of Science and Technology,
 $^{23}$Maseno University,
 $^{24}$Tonative Africa,
 $^{25}$University of Bath,
 $^{26}$National University Philippines,
 $^{27}$University of Copenhagen,
 $^{28}$Independent,
 $^{29}$University of Edinburgh,
 $^{30}$MIT-IBM Watson AI Research,
 $^{31}$MIT,
 $^{32}$IBM Research,
 $^{33}$Stanford University,
 $^{34}$Benha University,
 $^{35}$University of Hamburg,
 $^{36}$Bina Nusantara University,
 $^{37}$SEACrowd,
 $^{38}$Computational Intelligence and Operations Laboratory,
 $^{39}$University of New Haven,
 $^{40}$IT University of Copenhagen,
 $^{41}$Ofis Publik ar Brezhoneg,
 $^{42}$Johns Hopkins University,
 $^{43}$New York University,
 $^{44}$TUM,
 $^{45}$Stevens Institute of Technology,
 $^{46}$Capital One,
 $^{47}$University of North Carolina at Chapel Hill,
 $^{48}$Academia Aragonesa de la Lengua,
 $^{49}$Universidad de Zaragoza,
 $^{50}$Liverpool John Moores University,
 $^{51}$MBZUAI,
 $^{52}$DFKI Berlin,
 $^{53}$The African Research Collective,
 $^{54}$VISTEC,
 $^{55}$National Research Council, Canada,
 $^{56}$Princeton University,
 $^{57}$University of The People,
 $^{58}$Saarland University,
 $^{59}$Birla Institute of Technology and Science,
 $^{60}$LORIA,
 $^{61}$University of Mannheim,
 $^{62}$NEC Laboratories Europe, Germany,
 $^{63}$NeuralShift,
 $^{64}$University of Cambridge,
 $^{65}$Hasso Plattner Institute,
 $^{66}$ELLIS Unit Potsdam,
 $^{67}$Universitas Pelita Harapan,
 $^{68}$University of Manchester,
 $^{69}$Thammasat University,
 $^{70}$ScienciaLAB,
 $^{71}$Inria Paris,
 $^{72}$University of Bucharest,
 $^{73}$Monash University, Indonesia,
 $^{74}$Wadmedani Ahlia University,
 $^{75}$NSUK,
 $^{76}$Council for Ligurian Linguistic Heritage

\onecolumn

\section{CommonLID statistics by language}
\label{sec:appendix_stats}
\begin{table*}[h!]
    \centering
    \small
    \begin{tabular}{crrp{1.5cm}crr}
    \toprule
         \textbf{Lang. code} & \textbf{Num. lines} & \textbf{Mean length} & & \textbf{Lang. code} & \textbf{Num. lines} & \textbf{Mean length}  \\
         \cmidrule(lr){1-3} \cmidrule(lr){5-7}
            ace & 1 & 28.0 & & jpn & 3,344 & 171.9 \\
            acf & 603 & 241.0 & & kab & 116 & 118.2 \\
            aeb & 15 & 220.1 & & kan & 2,554 & 286.1 \\
            afr & 88 & 193.7 & & kik & 62 & 237.2 \\
            amh & 1,617 & 207.0 & & kor & 8 & 70.1 \\
            apd & 27 & 224.5 & & lat & 50 & 355.7 \\
            ara & 16,306 & 184.3 & & lav & 512 & 214.1 \\
            arb & 26,152 & 211.4 & & lij & 147 & 320.1 \\
            arg & 2,342 & 333.7 & & lin & 55 & 246.1 \\
            ars & 229 & 224.9 & & ltg & 38 & 234.5 \\
            ary & 226 & 180.8 & & lug & 1,361 & 170.6 \\
            arz & 1,102 & 243.5 & & lvs & 353 & 287.4 \\
            asm & 213 & 247.9 & & mal & 2,061 & 266.6 \\
            aze & 29 & 227.6 & & mar & 1,061 & 278.8 \\
            azj & 847 & 277.4 & & mlg & 2,153 & 197.0 \\
            bak & 46 & 136.0 & & msa & 28,224 & 188.8 \\
            bcl & 270 & 331.9 & & nld & 3,299 & 263.9 \\
            ben & 1,886 & 204.5 & & nso & 99 & 137.0 \\
            bik & 1,499 & 268.6 & & nyn & 5 & 284.4 \\
            bre & 2,348 & 183.3 & & oci & 1,314 & 225.6 \\
            bul & 109 & 190.9 & & orm & 1,060 & 233.7 \\
            cat & 93 & 262.6 & & ory & 608 & 197.0 \\
            ces & 933 & 275.0 & & pan & 1,020 & 234.9 \\
            cmn & 865 & 169.4 & & pcm & 2 & 20.5 \\
            crh & 405 & 235.2 & & pol & 1 & 273.0 \\
            deu & 7,553 & 218.7 & & por & 2,443 & 171.7 \\
            ell & 7 & 23.9 & & rcf & 401 & 195.7 \\
            eng & 27,461 & 212.9 & & rus & 4,003 & 220.9 \\
            est & 659 & 174.3 & & san & 895 & 282.3 \\
            ext & 7 & 91.7 & & sna & 1,355 & 161.1 \\
            fas & 19,318 & 237.0 & & sot & 943 & 455.6 \\
            fil & 58 & 229.1 & & spa & 4,236 & 212.7 \\
            fin & 1,030 & 267.3 & & swa & 4,031 & 158.7 \\
            fra & 3,233 & 206.7 & & swh & 12,383 & 222.5 \\
            fro & 39 & 740.9 & & tam & 81 & 218.5 \\
            fry & 965 & 265.7 & & tat & 1,029 & 203.3 \\
            fuv & 37 & 258.7 & & tel & 11,747 & 223.0 \\
            gaz & 11 & 226.2 & & tgl & 2,223 & 217.3 \\
            gcf & 24 & 157.9 & & tha & 3,118 & 237.0 \\
            gcr & 111 & 156.1 & & tuk & 22 & 285.4 \\
            gla & 929 & 202.6 & & tur & 4,486 & 233.7 \\
            gle & 1 & 23.0 & & ukr & 2 & 37.0 \\
            gom & 338 & 219.8 & & urd & 204 & 232.4 \\
            grc & 1 & 7.0 & & uzb & 43,189 & 206.9 \\
            gug & 548 & 219.2 & & uzs & 9 & 409.8 \\
            guj & 948 & 222.4 & & vec & 1,558 & 193.8 \\
            guw & 4 & 71.0 & & vie & 21,803 & 213.1 \\
            hau & 16,455 & 166.4 & & wuu & 631 & 137.9 \\
            hbo & 808 & 535.3 & & xho & 575 & 257.7 \\
            heb & 5,055 & 167.4 & & yor & 2,290 & 93.5 \\
            hin & 3,666 & 249.2 & & yue & 425 & 91.2 \\
            ibo & 168 & 309.0 & & zho & 11,738 & 152.7 \\
            ind & 33,828 & 278.8 & & zsm & 343 & 172.8 \\
            ita & 4,387 & 229.7 & & zul & 4 & 20.2 \\
            jav & 1,656 & 231.5 & & & & \\
         \bottomrule
         & 
    \end{tabular}
    \caption{Number of lines and mean line length in characters for each language variety in the CommonLID dataset.}
\end{table*}

\newpage

\section{Examples of CommonLID Errors} \label{sec:examples_errors}

\begin{table*}[ht]
\centering
\small
\begin{tabular}{p{0.6\linewidth} l l}
\toprule
\textbf{Text} & \textbf{Gold Label} & \textbf{GlotLID Label} \\
\midrule
Tags: Champions League, Liverpool F.C., Real Madrid C.F.
  & English & Banjar \\
Doklat: Bulghanghan ilahi hoquqlar: Xitayning Uyghur dini erkinligige qaratqan qattiq qol basturishliri
  & Uzbek & Uyghur \\
GABRIEL FAUR\'E Piano Quartet No.\ 1 in C Minor, Op.\ 15
  & English & Afrikaans \\
$\uparrow$ ``Boxing Records, Lb for Lb''. boxrec.com. Retrieved October 24, 2014.
  & English & Scots \\
\^{} ``Language Log --- Ask Language Log: How to pronounce `Antifa'?''. languagelog.ldc.upenn.edu.
  & English & Manx \\
SMS TUG'ILGAN KUN TABRIKLARI | \cyr{Просмотров: 21967 | Добавил: SheJoooT | Теги: SMS tug'ulgan kun tabriklari \ldots}
  & Uzbek & Hausa \\
\^{} 7.00--7.10 \zh{楊南郡、王素娥。《玉山國家公園八通關越嶺古道西段調查研究報告》(PDF)。玉山國家公園。1987。}
  & Chinese & Hebrew \\
Peugeot Paris 36249 -- Molinillo el\'ectrico de pimienta (34\,cm, \ldots)
  & Spanish & Finnish \\
Le magazine des sites Calvin Klein\textsuperscript{\textregistered} Established 1978
  & French & Northern Sotho \\
Conseil en investissement / Conseil en investissement 10 (Aube) / Mesnil-Saint-P\`ere
  & French & Dutch \\
\bottomrule
\end{tabular}
\caption{Examples from the CommonLID dataset which are misclassified by GlotLID.}
\end{table*}

\section{Additional Model Details} 
\label{app:model_details}

\begin{table*}[hbt]
    \centering
    \small
    \begin{tabular}{lcccc}
    \toprule
         Name & \# langs. & Architecture & Data sources & Open data? \\
         \midrule
         AfroLID \citep{adebara-etal-2022-afrolid} & 517 & Transformer & $\approx$ 100M curated sents. & \xmark{} \\
         CLD2 \citep{cld2} & 158 & Na\"{i}ve Bayes & Web pages (curated and  scraped) & \xmark{} \\
         fasttext \citep{nllb-24} & 218 & FastText & ``publicly available datasets'' & \xmark{} \\
         FUN-LangID \citep{caswell2024funlangid} & 1634 & Common sub-strings & Web+Wikipedia+Bibles & \xmark{} \\
         pyFranc \citep{pyfranc} & 414 & Trigram distribution & UDHR & \cmark{} \\
         gCLD3 \citep{cld3} & 99 & Neural network & \qmark{} & \xmark{} \\
         GlotLID v3 \citep{kargaran-etal-2023-glotlid} & 1868 & FastText & Curated open sources & \cmark \\
         OpenLID-v2 \citep{burchell-etal-2023-open} & 193 & FastText & Curated, audited open sources & \cmark{} \\
         \bottomrule
    \end{tabular}
    \caption{Summary of LID models used.}
    \label{tab:eval_models}
\end{table*}

\begin{table*}[htbp]
    \centering
    \small
    \begin{tabular}{lcccccccc}
    \toprule
        & GlotLID & FUN-LangID & AfroLID & pyFranc & fasttext & OpenLID-v2 & CLD2 & gCLD3 \\
        \midrule
        GlotLID & \textbf{1868} \\
        FUN-LangID & 1351 & \textbf{1549}\\
        AfroLID & 401 & 313 & \textbf{515}\\
        pyFranc & 362 & 312 & 93 & \textbf{410} \\
        fasttext & 202 & 180 & 49 & 166 & \textbf{210} \\
        OpenLID-v2 & 190 & 157 & 47 & 156 & 180 & \textbf{193} \\
        CLD2 & 128 & 153 & 30 & 121 & 114 & 104 & \textbf{158} \\
        gCLD3 & 83 & 99 & 13 & 80 & 79 & 78 & 98 & \textbf{99} \\
        \bottomrule
    \end{tabular}
    \caption{Counts of mutual language coverage between LID models (models in descending order by coverage).}
    \label{tab:coverage}
\end{table*}

\begin{table}[htb]
    \centering
    \small
    \begin{tabular}{lrr}
        \toprule
        Model & Samples/s & Duration @$1e+6$ samples \\
        \midrule
        CLD2 & 43735 & 22s \\
        OpenLID-v2 & 13123 & 1m 16s \\
        fasttext & 10867 & 1m 32s \\
        FUN-LangID & 7237 & 2m 18s \\
        GlotLID & 3127 & 5m 19s \\
        gCLD3 & \textsuperscript{$\ddagger$}2026 & 8m 13s \\
        Franc & 520 & 32m 03s \\
        AfroLID & \textsuperscript{$\dagger$}66 & 4h 12m 49s \\
        \bottomrule
    \end{tabular}
    \caption{Inference speed for LID models on FLORES+. Models are listed in descending order of samples per second. For all models other than gCLD3, inference is done on a 14-core Apple M4 Pro chip with 64GB of RAM, using PyTorch MPS optimisation where possible. \textsuperscript{$\dagger$}AfroLID uses transformers - performance is likely significantly better on CUDA hardware. \textsuperscript{$\ddagger$}Measured on an AMD EPYC 7351P (16 cores @ 2.4GHz) Linux machine with 256GB RAM as gCLD3 is not usable on modern macOS. }
    \label{tab:speed}
\end{table}

\begin{figure*}[ht!]
\begin{center}
\includegraphics[width=0.8\textwidth]{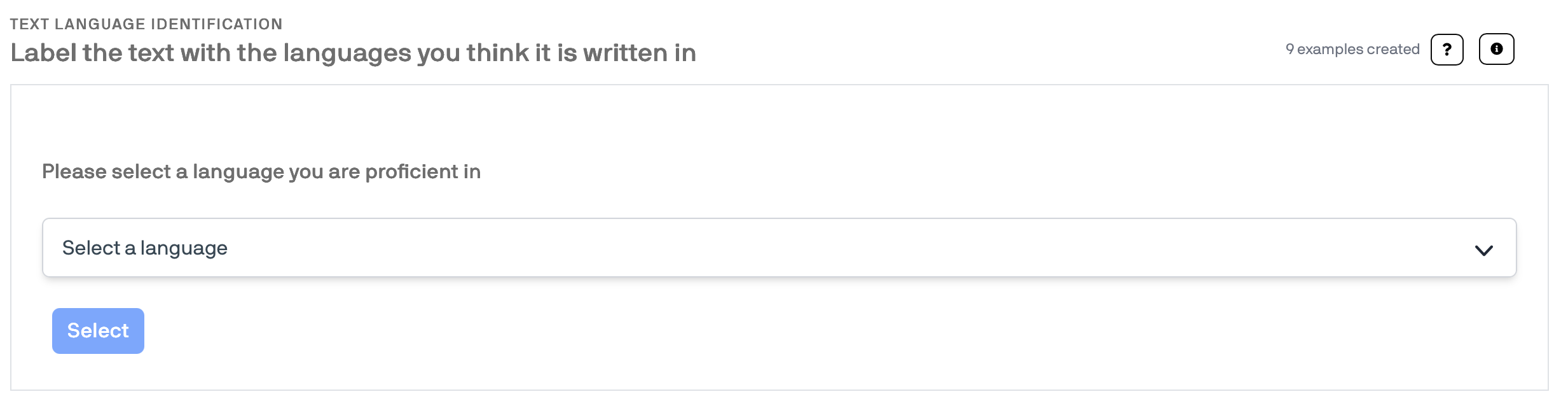}
\caption{A screenshot of the annotation platform interface. The participant selects the language they want to annotate for.}
\label{fig:platform_screenshot}
\end{center}
\end{figure*}

\twocolumn

\section{Annotation platform details}\label{app:annotation_platform}

\subsection{Annotation Instructions}

In this task, annotators will be first give a prompt in which they select a language that they are proficient on. The bar is a search field so that the annotator can easily find the language they are looking for, as can be seen in \Cref{fig:platform_screenshot}.

\begin{figure*}[h]
\begin{center}
\includegraphics[width=0.75\textwidth]{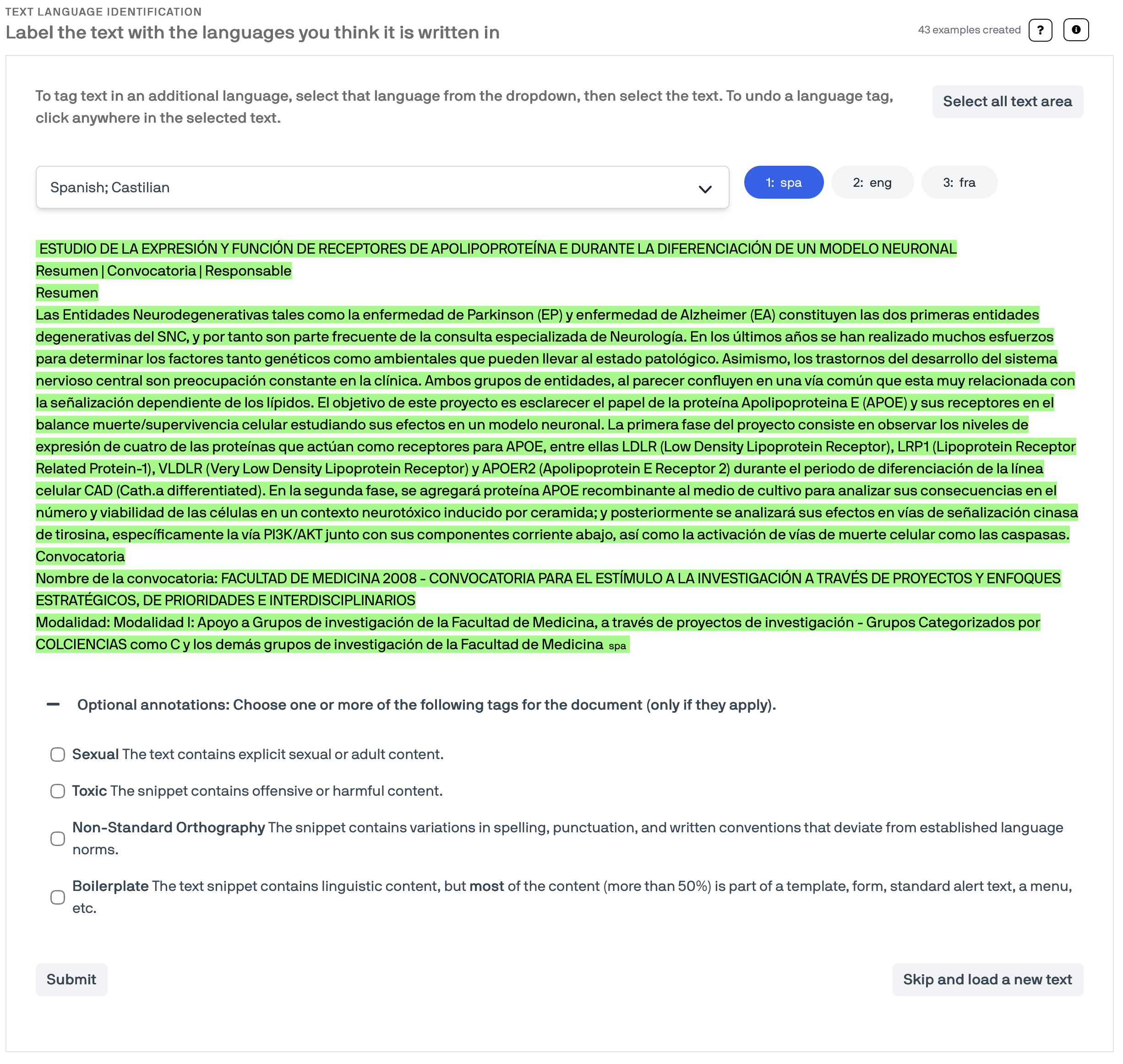}
\caption{A screenshot of the annotation platform interface. The participant has highlighted all text as Spanish.}
\label{fig:platform_screenshot2}
\end{center}
\end{figure*}

Then they will be presented with a text passage from a processed Common Crawl record that potentially contains content in the selected language. The annotator will then select the spans of text that they are able to identify as being written in the language they are proficient in. If the whole text passage is written in a single language, the annotator can just press the ``select all text area button'' to select all the text, as can be seen in \Cref{fig:platform_screenshot2}.

There are some optional annotations that the user might wish to add, they are displayed at the bottom of the image in a dropdown. These tags are optional and are intended to start preparing future tasks.

\begin{figure*}[h]
\begin{center}
\includegraphics[width=0.8\textwidth]{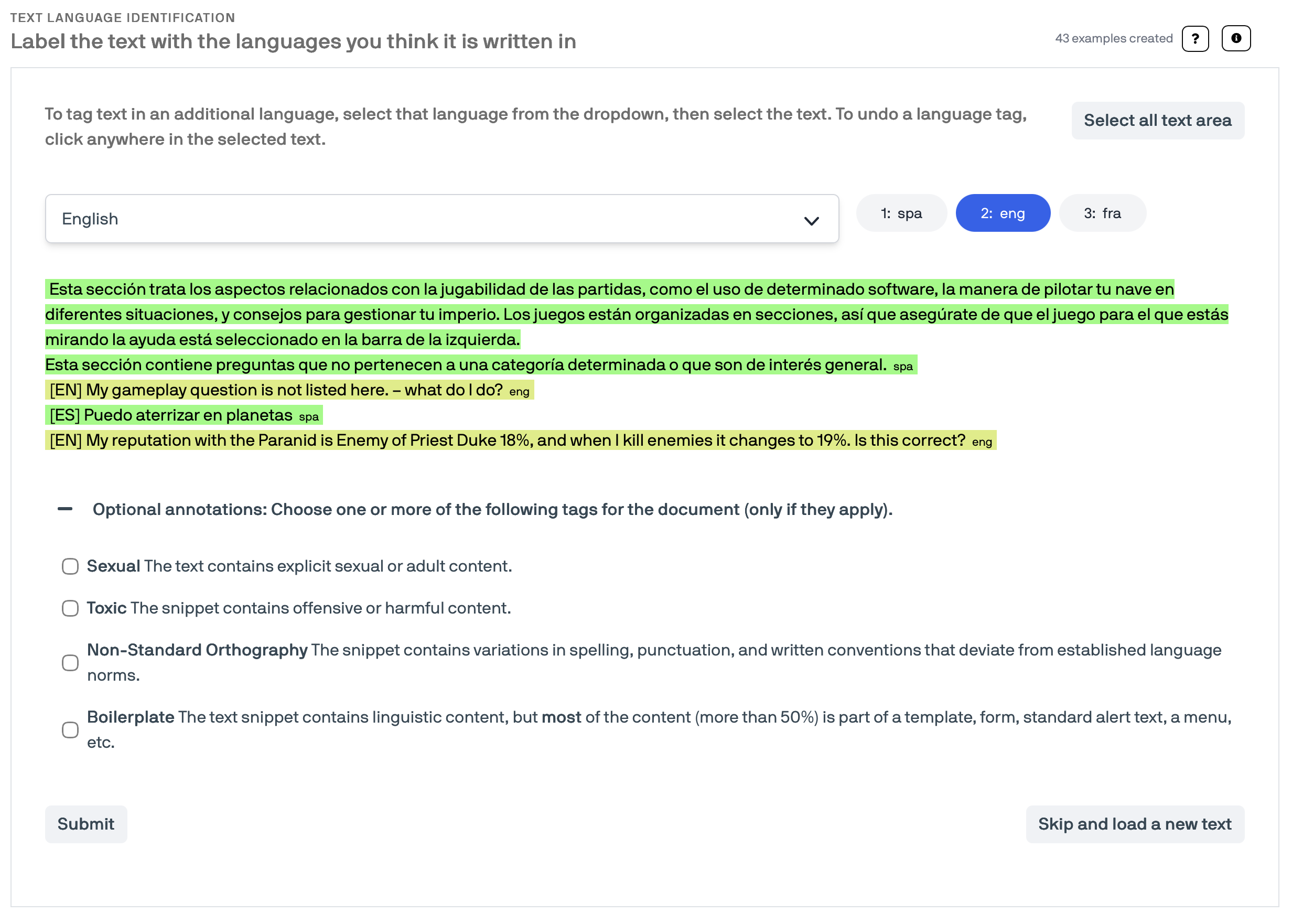}
\caption{A screenshot of the annotation platform interface. The participant has highlighted the English and Spanish text in the extract in different colours.}
\label{fig:platform_screenshot3}
\end{center}
\end{figure*}

If there are multiple languages in a single example and you can identify them, you can select an another language from the list and the select the span of text. Please try to annotate no more than single language per line. If you see any instance of code-switching or even script-switching in a single line, please try to annotate it with the language that you think is ``dominant'', as the example in \Cref{fig:platform_screenshot3}.

If at some point you wish to annotate examples in another language, simply select the desired language on the dropdown, and click on the ``skip and load new text'' button.

\subsection{Potential Issues}

Samples have been automatically pre-annotated by ML models so the performance for some languages might be lower than ideal. If you have seen a large amount of samples that are incorrectly labelled, please report it by clicking on the button labelled with an ``i'' on the to right corner. This will show you a pop-up with instructions on how to report issues with the data.

\subsection{Rewards for Participation}

All contributors who complete 100 annotations or more will be invited to be co-authors of a scientific paper.



\section{Evaluation Dataset Details} \label{app:eval_datasets}

\paragraph{FLORES+} This is a multilingual machine translation benchmark sourced from Wikitext articles \citep{nllb-24,guzman-etal-2019-flores,goyal-etal-2022-flores,maillard-etal-2024-findings,dale-etal-2025-findings}. We evaluated on version 4.1 of the dev split, which contains 997 parallel sentences translated into 222 language varieties. FLORES+ is a common \gls{lid} evaluation set due to its high quality and wide coverage, though it only contains relatively formal text. 

\paragraph{UHDR-LID} This is a collection of cleaned translations of the Universal Declaration of Human Rights into 374 languages \citep{kargaran-etal-2023-glotlid}. Whilst its high language coverage and open availability make it attractive as a \gls{lid} evaluation dataset, these factors also mean it is often used as part of multilingual training data. This means that scores on this data are likely to be inflated.

\paragraph{SmolSent} This dataset consists of 863 English sentences covering 5519 of the most common English tokens, professionally translated into 88 under-served languages \cite{caswell-etal-2025-smol}. These sentences come from Common Crawl and are chosen as the smallest set of sentences covering the largest number of common English tokens. Their vocabulary coverage and web domain makes them a useful evaluation set for our task.

\paragraph{Bible} We sourced translations of parts of the Bible in 1144 language varieties, based on the work of \citet{mayer-cysouw-2014-creating}. Different language varieties have varying amounts of data available, from 361,74 lines for English to just 5 for Ndj\'{e}bbanatha. Religious text, particularly the Bible, is often used as the sole source of both training and test data for many under-served language varieties, the result of which is that downstream models for this language only perform well on religious text and do not generalise to other domains.

\paragraph{Social media} To test on more informal text, we curate a multilingual dataset of 169,019 lines of social media data containing examples of posts in 97 language varieties. The largest language variety class contains 28,545 lines, whilst the seven smallest contain just one example each. This dataset contains phenomena typical of web text such as emoji, URLs, non-standard orthography and non-linguistic content like hashtags, making it a challenging but useful domain for testing robustness.



\end{document}